\documentclass[10pt,twocolumn,letterpaper]{article}
\usepackage{listings}

\usepackage{cvpr}
\usepackage{times}
\usepackage{epsfig}
\usepackage{graphicx}
\usepackage{svg}
\usepackage{amsmath}
\usepackage{amssymb}
\usepackage{multirow}
\usepackage{graphicx}
\usepackage[numbers]{natbib}
\usepackage{caption}
\usepackage{float}
\usepackage{changepage}
\usepackage{subfig}

\usepackage[colorlinks]{hyperref}

\cvprfinalcopy % *** Uncomment this line for the final submission

\def\cvprPaperID{****} % *** Enter the CVPR Paper ID here

\ifcvprfinal\fi
\begin{document}

\title{A Mixture of Experts Approach to 3D Human Motion Prediction}

\author{Edmund Shieh\thanks{Alphabetical order} , Joshua Lee Franco \footnotemark[1] , Kang Min Bae\footnotemark[1] , Tej Lalvani\footnotemark[1]\\
Georgia Institute of Technology\\
{\tt\small eshieh3@gatech.edu, jfranco33@gatech.edu, kbae36@gatech.edu, tlalvani3@gatech.edu}
}

\maketitle

\begin{abstract}
This project addresses the challenge of human motion prediction, a critical area for applications such as autonomous vehicle movement detection. Previous works have emphasized the need for low inference times to provide real time performance for applications like these \cite{shafaei2016realtime,EgoLocate2023,nargund2023spotr}. Our primary objective is to critically evaluate existing model architectures, identifying their advantages and opportunities for improvement by replicating the state-of-the-art (SOTA) Spatio-Temporal Transformer model as presented by \citet{aksan2020spatiotemporal} as best as possible given computational constraints. These models have surpassed the limitations of RNN-based models and have demonstrated the ability to generate plausible motion sequences over both short and long term horizons through the use of spatio-temporal representations. We also propose a novel architecture to address challenges of real time inference speed by incorporating a Mixture of Experts (MoE) block within the Spatial-Temporal (ST) attention layer, inspired by \citet{lepikhin2020gshard}. The particular variation that is used is Soft MoE \cite{puigcerver2023sparse}, a fully-differentiable sparse Transformer that has shown promising ability to enable larger model capacity at lower inference cost.

\end{abstract}

\section{Introduction}
Human motion prediction is a critical component in various applications such as autonomous vehicles and interactive robotics, where understanding and anticipating pedestrian movements can drastically improve safety and interaction dynamics. Traditional methods primarily utilize Recurrent Neural Networks (RNNs) due to their ability to process sequential information. However, these models often struggle with long-term dependencies and high inference times, which are crucial for real-time applications \cite{Martinez2017, Fragkiadaki2015}.

With the advancement of attention mechanisms, particularly the introduction of Transformers, new pathways have opened for handling sequential prediction tasks with improved accuracy and efficiency. The Spatial-Temporal (ST) Transformer model has emerged as a promising solution by effectively incorporating spatial and temporal information using a joint representation, which surpasses the limitations of RNN-based models in generating plausible motion sequences over varying horizons \cite{aksan2020spatiotemporal}.

Despite the successes of ST Transformers, real-time performance remains a challenge due to the the complexity and auto-regressive inference pattern of the models. To address this, we propose a novel architecture that incorporates a MoE within the ST Transformer's attention layers. This approach aims to optimize the inference speed by dynamically selecting the most relevant model components during the prediction process, drawing inspiration from recent advancements in scalable neural network technologies \cite{lepikhin2020gshard}. Current trends are gravitating towards increasing model complexity and expanding dataset sizes. This escalation necessitates enhanced computational power. Integrating a Mixture of Experts (MoE) with Spatial-Temporal (ST) Transformers could provide a strategic advantage by scaling model complexity while maintaining swift inference times. This approach aligns with the increasing demands for more sophisticated models, as MoE allows the model to selectively activate only the relevant parts of the network for specific tasks. By combining MoE with ST Transformers, which are adept at processing time-related data, the models can efficiently handle complex applications like human motion prediction. If successful, this integration could significantly benefit sectors such as autonomous vehicles, interactive robotics, virtual reality, and gaming, where rapid and accurate predictions are crucial for both safety and enhancing user experiences.

For our experiments, we utilized the AMASS (Archive of Mocap as Surface Shapes) DIP dataset hosted by the Max Planck Institute for Intelligent Systems \cite{dip2020project}, which consolidates various human motion capture datasets into a single repository, significantly simplifying the preprocessing and usage of mocap data for research purposes. Comprising 8,593 sequences, AMASS encompasses over 9 million frames sampled at 60 Hz translating to approximately 42 hours of recorded motion. Each joint in the dataset can be represented using different formats; for our study, we chose the axis angle (aa) representation, where each joint's orientation is described by a 3-dimensional vector \(E \in \mathbb{R}^3\) \cite{Martinez2017}.

We employed the fairmotion library \cite{gopinath2020fairmotion} for preprocessing the raw motion sequences into a structured format and used the dataset splits defined in the work \cite{Aksan2019StructuredPrediction} for training, validation, and testing of the motion prediction sequence modeling task. The task is approached as a sequence modeling problem, where the input consists of 120 consecutive poses, equivalent to two seconds of motion captured at a frequency of 60Hz. The output is defined as a sequence of 24 poses, representing 400 milliseconds of motion.

This report details our efforts to replicate the state-of-the-art ST Transformer model, assess its performance across diverse motion datasets, and introduce our novel MoE-enhanced ST Transformer model. Through rigorous testing and analysis, we aim to demonstrate the efficacy of our approach in reducing inference times while maintaining high accuracy in motion prediction.

\section{Approach}
Our approach was a three-step process. The initial step was to get the baseline performance from the existing models. Then compare the performance to the current state of the art. Finally, explore ways to enhance the current state of the art. 

There were five models provided by the Facebook research team \cite{gopinath2020fairmotion} in their motion prediction repository. seq2seq, basic seq2seq using encoder decoder; tied seq2seq, seq2seq with same LSTM used in both encoder and decoder; RNN, comprised of a single RNN; transfer encoder, which is a transformer-LSTM hybrid model; lastly, transformer. With these models, each team member trained a model using the default parameter to get our baseline.

After getting our baselines, we wanted to compare the performance between the existing models and the current state-of-the-art, ST Transformer \cite{aksan2020spatiotemporal}. In order to do this, we implemented it in PyTorch using a repository provided by the authors written for tensorflow as a foundation\cite{motion_transformer}. To get the optimal performance, we tuned several hyperparameters with various values to get the optimal model with the best settings. The hyperparameter that we tuned were batch size, optimizer, Dimensional joint embedding size, Hidden dimension, and number of layers.

Finally, taking inspiration from the promising results shown in \cite{lepikhin2020gshard}, we swapped out the dense feed forward layers within the transformer block of the vanilla ST transformer and replaced them with a Soft MoE block \cite{lucidrains2023st-moe-pytorch} and then compared the inference performance between the two under different hyperparameters. 

\subsection*{Joint Embedding}
The sequence of poses \( \mathbf{x} \) represents a tensor in \( \mathbb{R}^{T \times S \times M} \), where \( T \) is the number of frames in the input sequence (120 frames), \( S \) is the number of joints (24 joints), and \( M \) is the dimension of the axis angle representation for each joint (3 dimensions). This sequence is processed through a linear layer, which replaces the traditional embedding lookup table. This transformation outputs embeddings \( \mathbf{E} \in  \mathbb{R}^{T \times S \times E} \). Here, the input dimension \( M \) is transformed to the model's working dimension \( E \) an adjustable hyperparameter to tune complexity of the effective joint embedding.

\subsection*{Positional Encoding}
A Positional Encoding module is utilized to incorporate temporal information into the embeddings, enhancing the model's ability to understand sequence order. The positional encoding uses sinusoidal functions as proposed in section 3.5 of \citet{vaswani2017attention}, which are calculated according to equations \ref{eq:cosine_component} and \ref{eq:sine_component}.

\subsection*{Spatial and Temporal Multi-head Attention}

The core of the model lies in its dual attention mechanism as depicted in Figure 1 of \cite{aksan2020spatiotemporal}:
\begin{figure}[h] 
  \centering
  \includegraphics[width=0.5\textwidth]{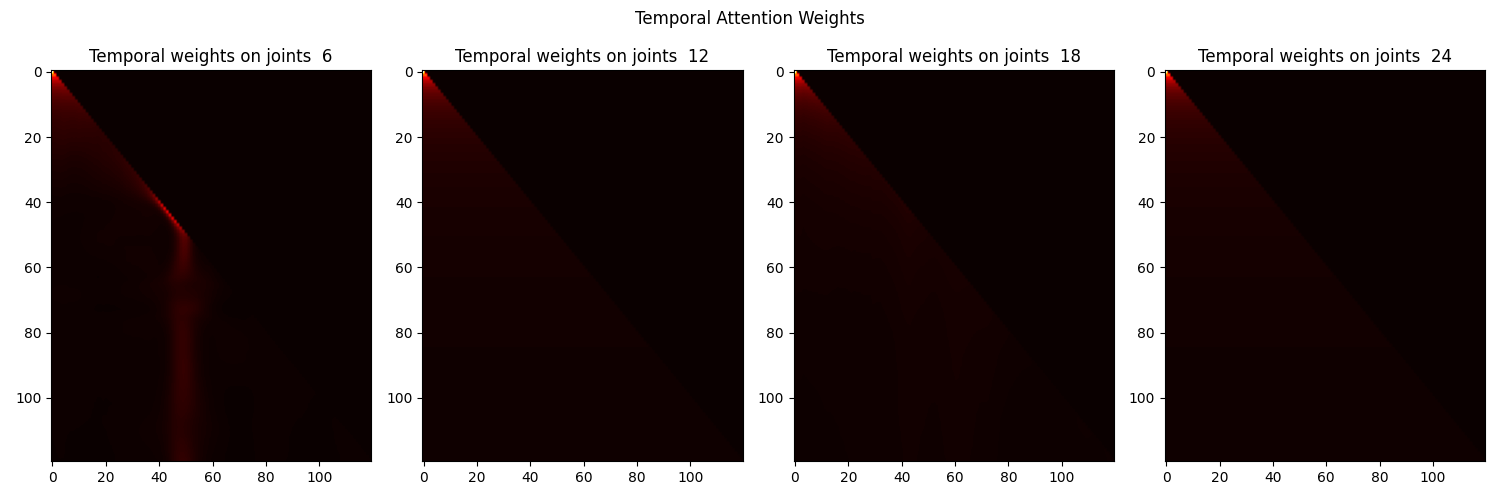} % ensure the filename matches and specify the desired width
  \caption{\textbf{Temporal Attention} map of temporal attention weights by timesteps in joints 6, 12, 18, 24}
  \label{fig:temporal_attention}
\end{figure}
\begin{figure}[h] 
  \centering
  \includegraphics[width=0.5\textwidth]{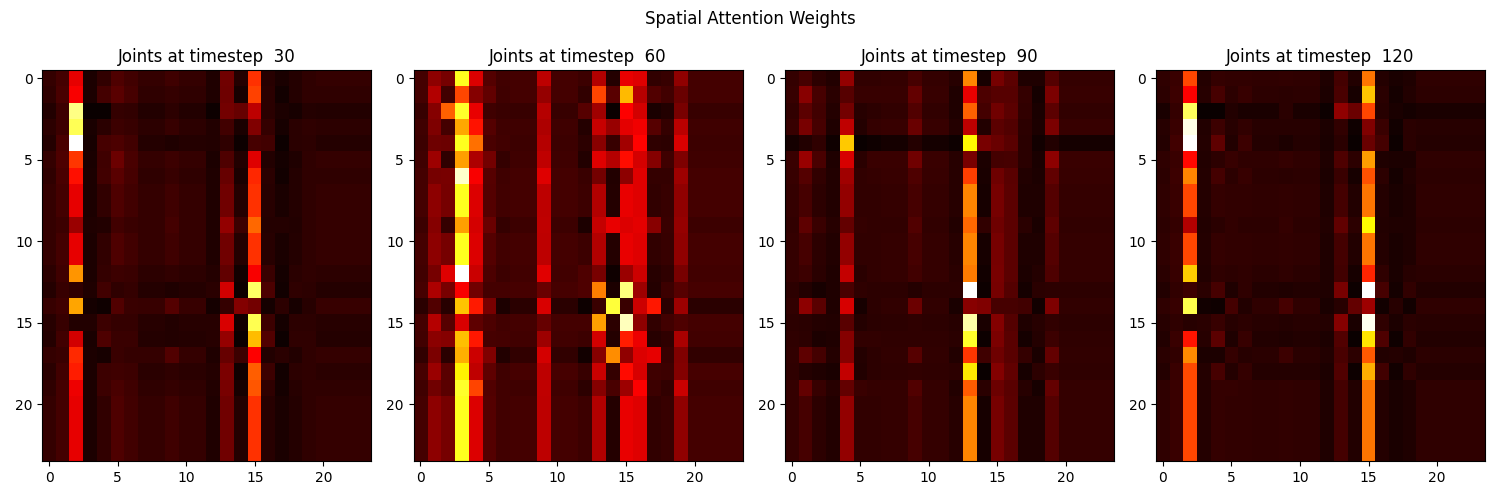} % ensure the filename matches and specify the desired width
  \caption{\textbf{Spatial Attention} map of spatial attention weights by joints 30, 60, 90, 120 timesteps}
  \label{fig:spatial_attention}
\end{figure}
\begin{enumerate}
    \item \textbf{Temporal Attention}:
    \begin{itemize}
        \item The input \( x \) is reshaped and transposed to align for temporal attention, resulting in a new shape of \( T \times (S \cdot E) \). This allows the multi-head attention layer to processes each joint independently across the time dimension, ensuring that each pose contributes to the final prediction without peeking at future poses.
        \item A masked multi-head attention is applied to incorporate causal dependency, allowing each time step to only attend to previous and current steps.
        \item Shown in figure \ref{fig:temporal_attention} is a map of self attention weight output of spatial multi head attention. It seems that the temporal attention patterns are similar across joints. In particular, most joints exhibit the temporal self-attention pattern in which the current timestep has notable attention weights with timesteps that just precede it ($<5$ frames back). This is demonstrated by a bright diagonal line on the heat maps. Furthermore, the upper right triangle is completely dark indicating successful causal masking in which a given timestep does not attend to future timesteps. Counter to our SOTA implementation, \citet{aksan2020spatiotemporal} SOTA model demonstrates different joints exhibiting different temporal attention patterns which are likely due to a more complex model being trained using higher number of heads in multi-head self attention and more layers allowing each head/layer to capture specific nuances for each joint.
    \end{itemize}

    \item \textbf{Spatial Attention}:
    \begin{itemize}
        \item The original input \( x \) is also reshaped to put spatial features at the forefront, resulting in a tensor shaped \( S \times (T \cdot E) \). This allows a separate multi-head attention layer to operate across all joints within a frame jointly attending to information from different joint embedding subspaces, allowing the model to capture the complex interdependencies between joints.
        \item Shown in figure \ref{fig:spatial_attention} is a map of self attention weight output of temporal multi head attention. This figure highlights that different joints exhibit different spatial attention patterns at different timesteps in line with the SOTA from \cite{aksan2020spatiotemporal}. We observe there are some important joints at each timestep that have high attention weights for all other joints. For example, joint 3 at timestep 60 has a bright column indicating it is attended to by most joints suggesting that this particular joint has high relevance to the model's prediction.
        
    \end{itemize}

\end{enumerate}

Both attention mechanisms are implemented using the \texttt{MultiheadAttention} PyTorch module which implement multi-head attention as described in \cite{vaswani2017attention}. The two attention outputs are then reshaped and added together. Note that the SOTA implements weight matrix sharing for key and value weights across joints whereas our implementation does not due no in-built support for this in \texttt{MultiheadAttention}.

\subsection*{Feedforward Layers (Vanilla ST Transformer)}
For the vanilla ST Transformer, each attention output is then processed through two sequential linear transformations interspersed with a ReLU activation function. The sequence is as follows:
\begin{align*}
    \text{Linear} &\rightarrow \text{ReLU} \rightarrow \text{Linear}.
\end{align*}
This feedforward network is applied independently to each position, followed by dropout and residual connections to stabilize the learning process.

\subsection*{Soft MoE layer (MoE ST Transformer)}
Sparse MoEs are shown to be able to scale model capacity in vision tasks without large increases in training or inference costs via processing inputs through a gating mechanism that selects which 'expert' networks to activate based on the input \cite{riquelme2021scaling}. The layer can be represented as:
\[
\mathbf{y} = \sum_{i=1}^n g_i(\mathbf{x}) \mathbf{E}_i(\mathbf{x}),
\]
where $g_i(\mathbf{x})$ are the gating functions determining the weights for each expert’s contribution, and $\mathbf{E}_i(\mathbf{x})$ represents the output from the $i$-th expert. The top $k$ out of $n$ experts are chosen to be active for any given input (where $k \ll n$). This reduces computational overhead significantly by activating only a small subset of the available experts, in contrast to dense layers that use all weights and biases for every input.

\begin{figure}[h] 
  \centering
  \includegraphics[width=0.5\textwidth]{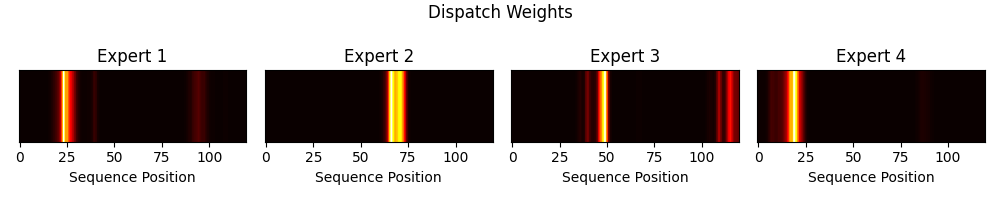} % ensure the filename matches and specify the desired width
  \caption{\textbf{MoE ST Transformer Dispatch Weights} Dispatch weights extracted from the MoE layer in a single forward pass of the trained MoE ST Transformer on an input sequence from the test split. The weights demonstrate a clear routing mechanism for the data to be processed by a specialized expert.}
  \label{fig:dispatch}
\end{figure}

Soft MoE \cite{puigcerver2023sparse}, an adaptation of the Sparse MoE, addresseses challenges like training instability and inability to scale the number of experts. Soft MoE performs an implicit soft assignment of input tokens to each expert. It utilizes slots which are essentially weighted averages of all input tokens, as determined by learned parameters. The slots are then processed by the experts, potentially lowering the computational cost even further while maintaining or enhancing model capacity and efficiency. This is achieved by computing weighted averages of all input tokens, where the weights are determined by the interaction between input tokens and a learned parameter matrix. This effectively results in each expert processing a subset of the data that it's specialized on as shown in \ref{fig:dispatch}, lowering computational cost.

\begin{figure*}[t] 
  \centering
  \includegraphics[width=.85\textwidth]{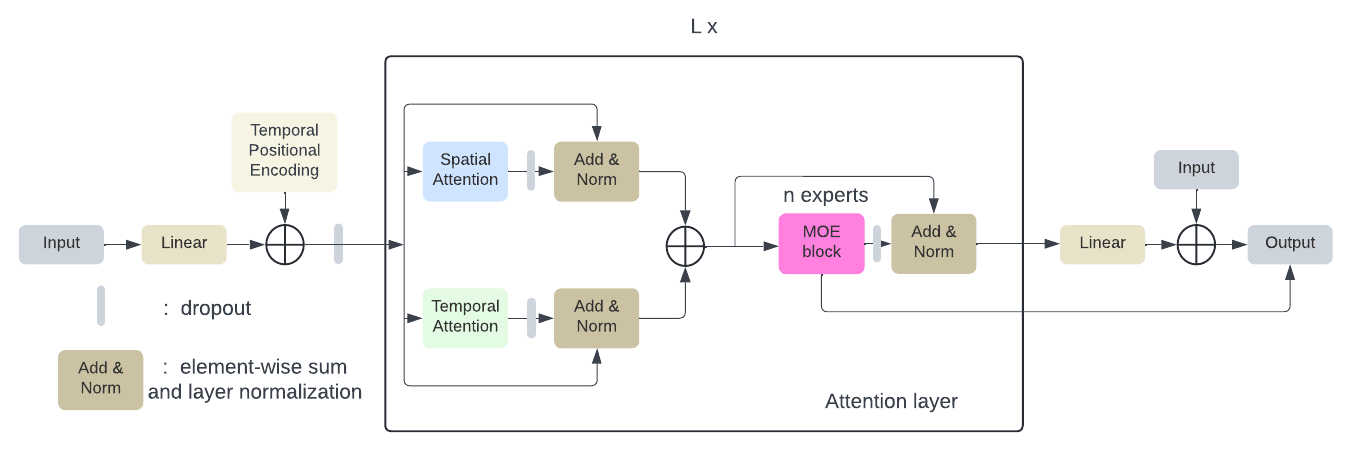} % ensure the filename matches and specify the desired width
  \caption{\textbf{Architecture Overview} Adapted version of ST transformer from Figure 2 in \cite{aksan2020spatiotemporal} with feed-forward layers in attention layers swapped with MoE block (highlighted in pink)}
  \label{fig:moe}
\end{figure*}

The Soft MoE layer was implemented using this repository \cite{lucidrains2023st-moe-pytorch} for the corresponding paper \cite{puigcerver2023sparse}. It was used in place of the feed-forward layer for the MoE ST Transformer as shown in Figure \ref{fig:moe}.

\subsection*{Normalization and Residual Connections}
Normalization is applied after each sub-layer (attention, feedforward and MoE), using \texttt{LayerNorm} module in PyTorch to stabilize the training dynamics. Residual connections are also used extensively to facilitate gradient flow and mitigate the vanishing gradient problem.

\subsection*{Projection Layers}
The output of the transformer is fed into two sequential projection layers to refine and reshape the neural network's output to the required dimensions for the task of next frame prediction. 
\begin{enumerate}
    \item \texttt{project\_1} Layer:
    \begin{itemize}
        \item The \texttt{project\_1} layer is primarily responsible for reducing the dimensionality of the encoder's output back to the original axis angle representation dimension size.
    \end{itemize}

    \item \texttt{project\_2} Layer:
    \begin{itemize}
        \item Following the initial projection, the \texttt{project\_2} layer further processes the data by mapping it to the final output dimensions required by the model which is a single frame, effectively collapsing the sequence length dimension of the tensor using learned weights from this linear layer.
    \end{itemize}

\end{enumerate}

\subsection*{Autoregressive Inference}
The model operates by maintaining a rolling window of the last 120 frames, which corresponds to 2 seconds of data, given the frame rate. This window is used to predict the next frame in the sequence. Mathematically, the prediction for the next frame can be expressed as:
\[
\textbf{x}_{t+1} = f(\textbf{x}_{t-119:t}),
\]
where \( \textbf{x}_{t-119:t} \) denotes the frames from \( t-119 \) to \( t \), and \( f \) represents the predictive function of the model.

Once the next frame is predicted, the model updates its frame window by discarding the oldest frame and including the newly predicted frame. This shift allows the model to continuously predict new frames based on the most recent data, including its own predictions. The model repeats this process to predict a total of 24 frames, corresponding to 400 milliseconds. This sequence of predictions is generated autoregressively, with each new frame prediction incorporating the most recent frame into the window.

\subsection{Challenges}

During the implementation of the spatio-temporal transformer model, several challenges were encountered which required specific solutions to ensure the efficiency and stability of the training process. Below are the main points:

\begin{itemize}
    \item \textbf{Numerical Stability:} We observed that the loss occasionally collapsed to zero, leading to NaN values during training. This was potentially due to exploding gradients. To address this issue:
    \begin{itemize}
        \item We modified the teacher forcing ratio calculation by adding a small dummy value to prevent it from reaching zero, thus ensuring numerical stability.
    \end{itemize}

    \item \textbf{Computational Constraints:} The training of our model was limited by the computational constraints imposed by Partnership for an Advanced Computing Environment (PACE) clusters, which includes a maximum of 512 CPU hours and 16 GPU hours per job, with a maximum walltime of 18 hours for CPU jobs and 16 hours for GPU jobs. Given this and high queue times for GPUs:
    \begin{itemize}
        \item We developed a functionality within our training script to automatically save progress and resume training from the last saved checkpoint. This ensured continuity in model training despite time constraints.
        \item Additionally, GPU memory emerged as a significant constraint, particularly for large models with a high number of parameters, such as the MoE ST transformer. To mitigate this, we employed a strategy of using float32 rather than double precision, despite knowing that it could result in worse performance. This trade-off was necessary to fit the model within the available memory constraints while still achieving reasonable training times. Additionally, we were unable to replicate the high parameter count of the SOTA using the same number of multi attention heads, layers and dimension sizes given these memory constraints and so a slimmer model was chosen instead.
    \end{itemize}

    \item \textbf{Handling Large Datasets:} Due to the large dataset sizes and the model's complexity, the expected training times were significantly lengthy, risking frequent interruptions. To mitigate this:
    \begin{itemize}
        \item We employed PyTorch's DataDistributedParallel module along with Hugging Face's Accelerate API to facilitate multi-node and multi-GPU training, significantly enhancing computational efficiency and reducing overall wall clock time.
        \item We also enhanced the existing code with a load function to continue from the last saved epoch. This addressed the issue of interruptions due to reaching the maximum wall time.
    \end{itemize}
\end{itemize}

\section{Experiments and Results}

\subsection{Success Criteria}
We had two primary goals: firstly, to successfully implement the state-of-the-art ST Transformer, as described in \cite{aksan2020spatiotemporal}, with appropriate hyperparameter tuning to surpass baseline models. Secondly, to implement a novel architecture, the MoE ST Transformer, using \cite{lucidrains2023st-moe-pytorch}, to compare its performance against the current state of the art and assess any gains in inference efficiency. Both goals were achieved, with details provided below.

\subsection{Experiment 1}
\subsubsection{Base Model Performance Analysis}
Our team successfully trained the base models provided by \cite{gopinath2020fairmotion} using the AMASS dataset \cite{AMASS:2019} consisting of more than forty hours of motion captured data. We used the default hyperparameter values consisting of values shown in table \ref{tab:default_hyperparam_values} and the results are shown in table \ref{tab:model_performance_table}. We were surprised to find that the seq2seq model outperformed transformer and transformer encoder models in all criterias including MAE, training loss and validation loss. This may indicate that the default hyper-parameter values in \ref{tab:default_hyperparam_values} better suits the seq2seq model than the transformer models. But it may be worth noting that training time on A100 for transformers were averaging about 2 minutes per epoch whereas seq2seq training on two A100s took about 4 minutes per epoch as expected.
\vspace{-.15in}
\subsubsection{Hyper-Parameter tuning for ST Transformer}
The hyper parameter tuning results can be seen in \ref{tab:hyperparam_table}. Default values were 64 for joint embedding dimension; 1 for number of layers; 120 for source length; 64 for batch size; adam for optimizer; twenty epochs. Then for each hyper parameter, we adjusted the value while leaving all else as default values to measure the impact of change in given hyper parameter value. The best hyperparameter values can be seen on table \ref{tab:best_hyperparams_table}. For some of the hyper parameters such as hidden dimension like batch size, source length, and hidden layers, we chose broad ranges as we wanted to get an idea if increasing or decreasing these values would lead to a significant increase in performance. For values like joint embedding dimension size and number of layers were tuned in small ranges to capture narrower performance comparisons. We also observed that hyperparameters such as joint dimension sizes and hidden dimensions require higher GPU ram as they increased. Which is why we did not increase these values above certain limit. After hyperparameter tuning we observed that the ST transformer outperformed most of the base models as shown on table\ref{tab:model_performance_table} as a whole as it had the lowest training loss and validation loss compared to seq2seq although they were similar in MAE values. Thus we ruled this experiment to be a success.

\subsection{Experiment 2}
\subsubsection{ST Transformer vs MoE Inference Time Ablation} \label{inference_time}
Table \ref{tab:inference} presents the run times observed for each configuration during inference. The parameter varied for the ST Transformer was the hidden dimension of the feedforward layer that follows the spatio and temporal attention layers in the model. For the MoE ST Transformer which has these feedforward layers replaced by an MoE block, the parameter varied is the number of experts. All other parts of the architecture and computing resources remained the same. 

\begin{figure}[ht] 
  \centering
  \includegraphics[width=0.5\textwidth]{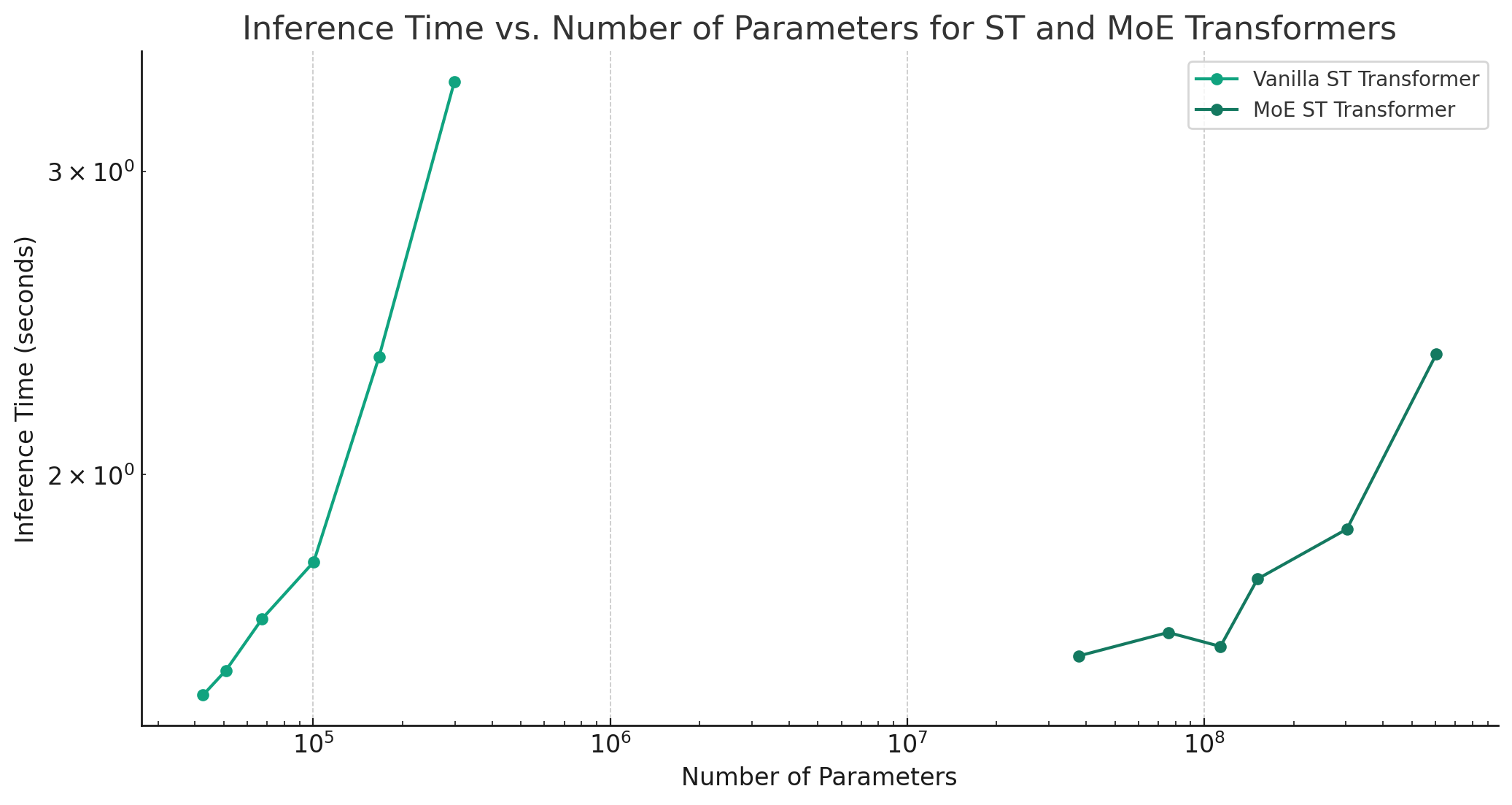} % ensure the filename matches and specify the desired width
  \caption{Graphical visualization of table \ref{tab:inference} to demonstrate scalability of MoE and ability to handle large number of parameters}
  \label{fig:inference}
\end{figure}

We observe in figure \ref{fig:inference} the ability for the MoE ST transformer to scale it's number of parameters without the same proportional increase in inference time compared to the Vanilla ST Transformer. A single Nvidia Quadro Pro RTX6000 GPU was used for this experiment. We use one slot per expert when scaling the MoE ST transformer as optimally recommended by \citet{puigcerver2023sparse}.

Traditional neural network models, such as the STtransformer, exhibit increased computational complexity and longer inference times as the hidden dimension size expands. This scaling effect is standard in neural networks where larger models require more computation per input, directly impacting performance metrics such as inference time.

The MoE model introduces a different architectural approach where only a subset of the model, termed "experts," is active at any given time. This design allows the model to scale in complexity, such as increasing the number of experts, without a corresponding linear increase in computational demand. As a result, inference times remain relatively stable, making the MoE model more scaleable and efficient for larger configurations.

Central to the SoftMoE architecture are two distinct types of weights, each serving a crucial role within the model: dispatch weights and combine weights. These weights are integral to the model's scalability and efficiency:

\begin{enumerate}
    \item \textbf{Dispatch Weights:} Responsible for routing input sequences to the appropriate experts, dispatch weights are determined by a gating network that evaluates each part of the input sequence. The weights decide which expert should process which frames within a sequence, ensuring that experts are effectively specialized to handle phases of the input sequence as seen in figure \ref{fig:dispatch}.
    
    \item \textbf{Combine Weights:} After the input has been processed by the designated experts, combine weights are used to aggregate the outputs from these experts. These weights determine the contribution of each expert's output to the final prediction, allowing for an effective synthesis of expert advice to achieve the most accurate overall output.
    
\end{enumerate}

Through these, the model dynamically adapts computational load across different experts, thereby maintaining efficiency despite increases in model parameters or complexity.

\begin{figure}[ht] 
  \centering
  \includegraphics[width=0.45\textwidth]{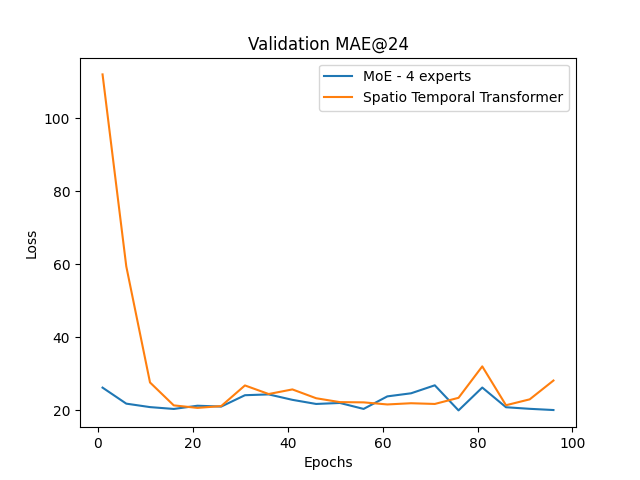} % ensure the filename matches and specify the desired width
  \caption{Validation MAE@24 of ST and MoE Transformers}
  \label{fig:MoeVsSTT}
\end{figure}

We observe in figure \ref{fig:MoeVsSTT} that the ST and MoE Transformers have similar performance on the validation set. This demonstrates how the MoE Transformer is able to scale without a reduction in performance.

\subsection{ST Transformer vs MoE evaluation}
We can see overall that MoE was able to perform well against the current state of the art ST Transformer. In table \ref{tab:test_model_performance}, both models outperformed the base models. This is to be expected, as the ST transformer incorporates both spatial and temporal attention as described above. This allows the model to better capture patterns in the motion data. The effectiveness of these models can be further seen in figures \ref{fig:gt1}, \ref{fig:pm1}, \ref{fig:gt2}, and \ref{fig:pm2}. In these figures, we see a ground truth motion compared to a predicted motion. From the perspective of joint distance, both models come close to predicting the reference motion given an input sequence preceding the motion. There is still room for improvement on this approach, though a promising area of improvement that MoE showed over the ST Transformer was in scalability of hyper-parameters and its affect on inference timing discussed in \ref{inference_time}.

\vfill
\clearpage
\bibliographystyle{unsrtnat}
\bibliography{bib}
\clearpage
\appendix

\section{Project Code Repository}

The GitHub repository available at \url{https://github.com/edshieh/motionprediction} extends the open-source \texttt{fairmotion} library, which provides foundational scripts and baseline models for motion prediction. The repository introduces several significant enhancements to facilitate more advanced and flexible research in motion prediction technologies. Key developments include:

\begin{itemize}
    \item Implementation of a PyTorch-based spatio-temporal (ST) transformer, which allows for advanced modeling of temporal dynamics and spatial relationships.
    \item Introduction of a mixture of experts within the ST transformer framework, enhancing the model's ability to handle diverse data scenarios by leveraging specialized sub-models.
    \item Enhanced training flexibility by allowing adjustments to hyperparameters such as the number of heads (\texttt{num\_heads}), layers (\texttt{num\_layers}), and experts (\texttt{num\_experts}) in the ST transformer.
    \item Improved inspection capabilities for attention and mixture of experts weights, providing deeper insights into the model's decision-making process.
    \item Addition of experimentation scripts designed to test inference times, aiding in the evaluation of model efficiency under different computational constraints.
\end{itemize}

\section{Model Specifics}
\paragraph{Positional Encoding:}
\begin{align}
    PE(\text{pos}, 2i) &= \sin\left(\frac{\text{pos}}{10000^{2i/d_{\text{model}}}}\right), \label{eq:sine_component} \\
    PE(\text{pos}, 2i+1) &= \cos\left(\frac{\text{pos}}{10000^{2i/d_{\text{model}}}}\right), \label{eq:cosine_component}
\end{align}
Where $PE(\text{pos}, 2i)$ and $PE(\text{pos}, 2i+1)$ represent the sine and cosine components of position encoding in a transformer model, respectively.

\begin{figure}[ht] 
  \centering
  \includegraphics[width=0.5\textwidth]{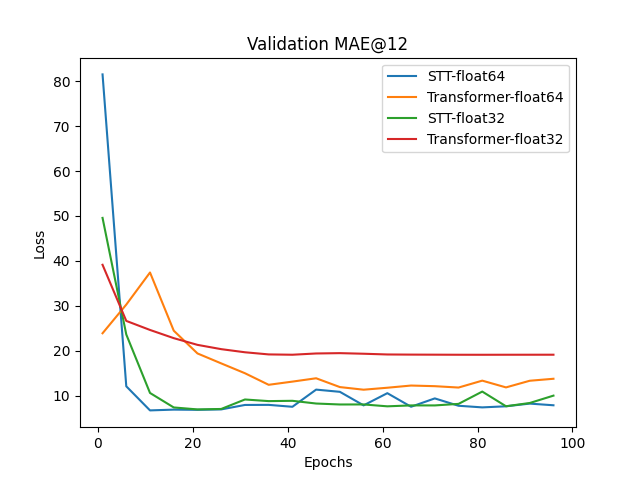} % ensure the filename matches and specify the desired width
  \caption{Comparison of float32 vs. float64 using validation MAE@12}
  \label{fig:FvD}
\end{figure}

\paragraph{Datatype Precision:} \label{datatype_precision}
We experimented with both float32 and float64 data-types for training different architectures. Across the board, we found that while training with float64 had a larger memory overhead and a longer training time, the validation MAE was lower than an identical model trained with float32. As seen in figure \ref{fig:FvD}, this makes a noticeable difference for different architectures. This is likely due to numerical stability of float64 given its higher precision, which ultimately leads to more precise gradients. In the interest of time, we trained all our models using float32. If performance was our goal, we would have opted to use float64. 

\section{Training Details}
\paragraph{Learning Rate Equation:}
The learning rate schedule is adapted to the training progress and is calculated using the following equation \cite{vaswani2017attention}:
\begin{equation}
    \text{learning rate} = D^{-0.5} \cdot \min\left(step^{-0.5}, step \cdot warmup^{-1.5}\right),
\end{equation}
where $D$ is the dimensionality of the model, $step$ denotes the current training step, and $warmup$ is the number of steps during the warm-up phase. This adaptive learning rate helps in stabilizing the training in the initial phases and gradually fine-tuning the model as training progresses. The default parameters were used from the noamopt option from the fairmotion library \cite{gopinath2020fairmotion}.

\paragraph{Loss Function for 3D Motion Prediction:}
The Mean Squared Error (MSE) loss was used as the loss criterion. Mathematically, the MSE loss for a prediction involving multiple joints across several time steps is defined as follows:

\begin{equation}
    L = \frac{1}{N \cdot J \cdot 3} \sum_{t=1}^{N} \sum_{j=1}^{J} \sum_{k=1}^{3} \left(\theta_{\text{tgt}, t, j, k} - \theta_{\text{pred}, t, j, k}\right)^2,
\end{equation}

where $N = 24$ represents the sequence v length, $J = 24$ denotes the number of joints, and $k$ indexes the three components of the axis-angle representation of each joint. Here, $\theta_{\text{tgt}, t, j, k}$ and $\theta_{\text{pred}, t, j, k}$ are the target and predicted values, respectively.

\paragraph{Error Metric for Assessing 3D Motion Prediction:}
Mean angle error (MAE) was used as the error metric to calculate the deviation between predicted and target pose. The mathematical formulation for the Euler angle error is as follows:

\begin{equation}
    E = \sqrt{\sum_{i \in \mathcal{I}} \left(\text{Euler}(R_{\text{tgt},i}) - \text{Euler}(R_{\text{pred},i})\right)^2},
\end{equation}
where $R_{\text{tgt},i}$ and $R_{\text{pred},i}$ are the target and predicted rotation matrices for joint $i$, respectively, and $\text{Euler}(\cdot)$ converts a rotation matrix to its Euler angle representation. The index set $\mathcal{I}$ includes only those joints for which the standard deviation of the target Euler angles exceeds a small threshold, indicating significant motion and thus relevance for the prediction accuracy.

\paragraph{Dynamic Teacher Forcing Ratio:}
\citet{pavllo2018quaternet} suggests to use teacher forcing to expose the model to its own predictions to mitigate exposure bias, which occurs when there is a discrepancy between the training regime (where the model always sees the true previous output) and inference (where the model only has access to its own predictions). By gradually exposing the model to its own predictions during training, teacher forcing helps the model better handle the sequential dependencies and learn more robust representations. During the forward pass of the network, at each timestep, the decision to apply teacher forcing is probabilistic, governed by the current value of the teacher forcing ratio. Conceptually, this can be represented as:
\[
    \text{Input at } t = 
    \begin{cases} 
    \text{True Output at } t-1 & \text{with p=TFR} \\
    \text{Model's Prediction at } t-1 & \text{otherwise}
    \end{cases}
\]

The degree to which teacher forcing as recommended by  is used is controlled by the \emph{teacher forcing ratio}, which is a function of the training progress:
\begin{equation}
    \text{Teacher Forcing Ratio} = \max\left(0, 1 - \frac{\text{current epoch}}{\text{total effective epochs}}\right)
\end{equation}
This ratio determines the probability at each timestep of using the true previous output rather than the model's prediction. As training progresses and the model becomes more capable of generating accurate predictions, the ratio decreases, gradually reducing the model's reliance on the true output. This transition helps the model to learn to generate sequences independently, improving its robustness and ability to generalize from training to inference scenarios.

\clearpage
\section{Hyperparameter Tuning}
\begin{table}[H]
    \centering  % Centers the table
    \begin{tabular}{|c|c| c | c |  c | c | c | c | c | c | c | c |}
         \hline
         \multicolumn{12}{|c|}{Hyperparameters and their values 20 epochs using Spatio Temporal Transformer} \\
         \hline
         J.dim & layers & Src Len & Batch Size & Optim & Hidden & T. Loss & V. Loss & MAE 6 & MAE 12 & MAE 18 & MAE 24 \\
         \hline
         16&1&120&64&adam&128&0.00185&0.00189&2.7055&7.4349&13.7244&20.9653 \\
         32&1&120&64&adam&128&0.00185&0.00189&2.7026&7.4326&13.7166&20.9568 \\
         48&1&120&64&adam&128&0.00185&0.00189&2.7022&7.4297&13.7132&20.9528 \\
         64&1&120&64&adam&128&0.00185&0.00189&2.7049&7.4341&13.7206&20.9629 \\
         80&1&120&64&adam&128&0.00185&0.00189&2.7021&7.4296&13.7131&20.9530 \\
         96&1&120&64&adam&128&0.00185&0.00189&2.7051&7.4339&13.7210&20.9614 \\
         112&1&120&64&adam&128&0.00185&0.00189&2.7025&7.4307&13.7147&20.9557 \\
         128&1&120&64&adam&128&0.00185&0.00189&2.7019&7.4275&13.7093&20.9491 \\
         64&2&120&64&adam&128&0.00185&0.00189&2.6861&7.4047&13.6753&20.9038 \\
         64&3&120&64&adam&128&0.00185&0.00189&2.6995&7.4244&13.7038&20.9436 \\
         64&4&120&64&adam&128&0.00185&0.00189&2.6991&7.4296&13.7015&20.9379 \\
         64&5&120&64&adam&128&0.00185&0.00189&2.7020&7.4275&13.7107&20.9509 \\
         64&1&120&32&adam&128&0.00367&0.00384&2.5957&7.2480&13.4984&20.7097 \\
         64&1&120&96&adam&128&0.00123&0.00125&2.7976&7.6433&14.0505&21.4340 \\
         64&1&120&64&sgd&128&0.00181&0.00191&2.8564&7.78822&14.24942&21.57944 \\
         64&1&120&64&noamopt&128&0.00059&0.00174&2.2324&6.9032& 13.3162&20.6464 \\
         64&1&90&64&noamopt&128&0.00229&0.00211&3.0059&8.6208&15.5506&23.1432 \\
         64&1&60&64&noamopt&128&0.00348&0.00234&3.1276&9.1852&16.1807&23.5763 \\
         64&1&30&64&noamopt&128&0.00183&0.00190&2.3958&7.4460& 13.9356&21.2564 \\
         64&1&120&64&adam&256&0.00185&0.00189&2.7048& 7.4330&13.7194&20.9617 \\
         64&1&120&64&adam&512&0.00184&0.00188&2.748& 7.5365&13.9419&20.3238 \\
         \hline
    \end{tabular}
    \caption{Table for Joint Dimensions, Number of Layers, Sequence Length, Batch Size, Optimizer, Hidden Dimension, Training Loss, Validation Loss, and MAE for 20 Epochs using Spatio Temporal Transformer}
    \label{tab:hyperparam_table}
\end{table}
\clearpage
\section{Model Performance}
\begin{table}[H]
\begin{center}
    \begin{tabular}{ | c | c | c | c | c | c | c | c | } 
         \hline
         \multicolumn{7}{|c|}{Base model, MoE, and ST Transformer performance with default values and 95 epochs} \\
         \hline
         Model & Training Loss & Validation Loss & MAE 6 & MAE 12 & MAE 18 & MAE 24 \\
         \hline
         RNN & 0.00593 &0.00437& 7.70777&15.92701&24.69410&33.78859 \\ 
         seq2seq & 0.00238 & 0.00164 & 2.82729 & 7.51121 & 13.78625 & 21.23050 \\
         transformer & 0.00427 & 0.00601 & 7.62462 & 19.089649 & 31.30014 & 43.94872 \\
         transformer encoder & 0.00702 & 0.00686 & 10.04142 & 20.43901 & 31.18811 & 42.193925 \\
         ST transformer & 0.00058 & 0.00050 & 2.78517 & 7.88353 & 14.46974 & 21.84928 \\
         MoE & 0.00346 & 0.00188 & 2.63867 & 7.90675 & 14.87096 & 22.81373 \\
         \hline
    \end{tabular}
    \caption{Base model, MoE, and ST Transformer performance with default values and 95 epochs}
    \label{tab:model_performance_table}
\end{center}
\end{table}

\begin{table}[H]
\begin{center}
    \begin{tabular}{ |c|c|c|c|c| } 
         \hline
         \multicolumn{5}{|c|}{ Default hyper-parameter values } \\
         \hline
         Batch Size&Optimizer&Joint Embedding Size&Hidden Dims&Num layers\\
         \hline
         64 & sgd & 56 & 1024 & 2 \\
         \hline
    \end{tabular}
    \caption{ Default hyper-parameter values used for base model training }
    \label{tab:default_hyperparam_values}
\end{center}
\end{table}

\begin{table}[H]
\begin{center}
    \begin{tabular}{ | c | c | c | c | c | }
         \hline
         \multicolumn{5}{|c|}{Hyperparameters and their optimal values} \\
         \hline
         Batch Size&Optimizer&Joint Embedding Size&Hidden Dims&Num layers\\
         \hline
         32 & noamopt & 128 & 512 & 2 \\
         \hline
    \end{tabular}
    \caption{ Best hyper parameter values for Batch size, Optimizer, Joint Dimensional Embedding Size, Hidden Dimensions, and Number of layers}
    \label{tab:best_hyperparams_table}
\end{center}
\end{table}

\begin{table}[H]
\begin{center}
    \begin{tabular}{ | c | c | c | c | c | c | c | c | } 
         \hline
         \multicolumn{7}{|c|}{ST Transformer vs MoE with Double/Float64 precision and 30 epochs} \\
         \hline
         Model & Training Loss & Validation Loss & MAE 6 & MAE 12 & MAE 18 & MAE 24 \\
         \hline
         ST transformer & 0.00408 & 0.00344 & 2.21965 & 6.80977 & 13.04387 & 20.19733 \\
         MoE & 0.00623 &  0.00359 & 1.90865 & 6.48504 & 12.98277 & 20.49475 \\
         \hline
    \end{tabular}
    \caption{ST Transformer vs MoE performance with using Double/Float64 precision on data types and and 30 epochs}
    \label{tab:float_vs_double_performance}
\end{center}
\end{table}

\clearpage
\begin{figure*}[h] 
\centering
\includegraphics[width=0.75\textwidth]{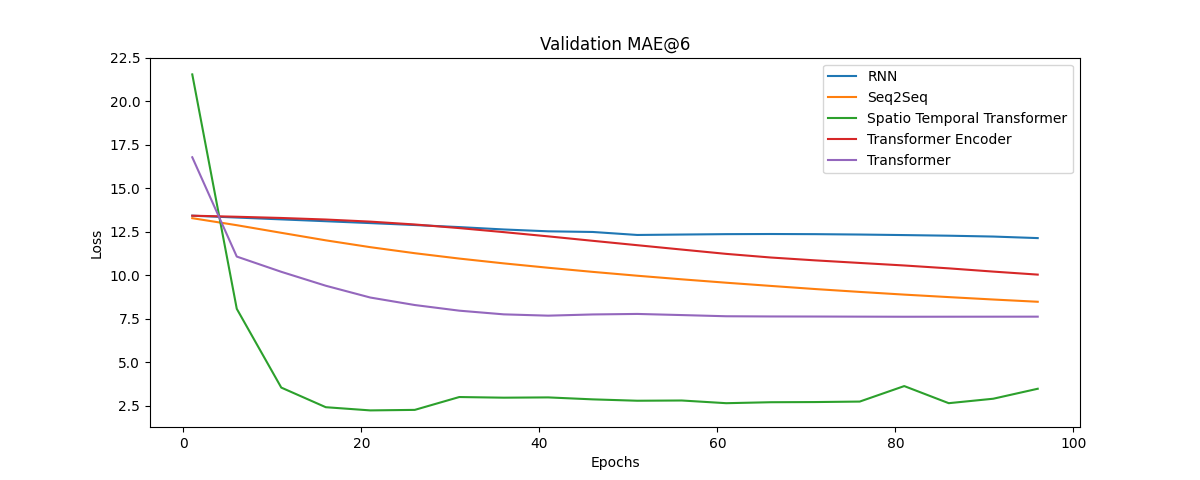}
\caption{\textbf Validation MAE@6}
\includegraphics[width=0.75\textwidth]{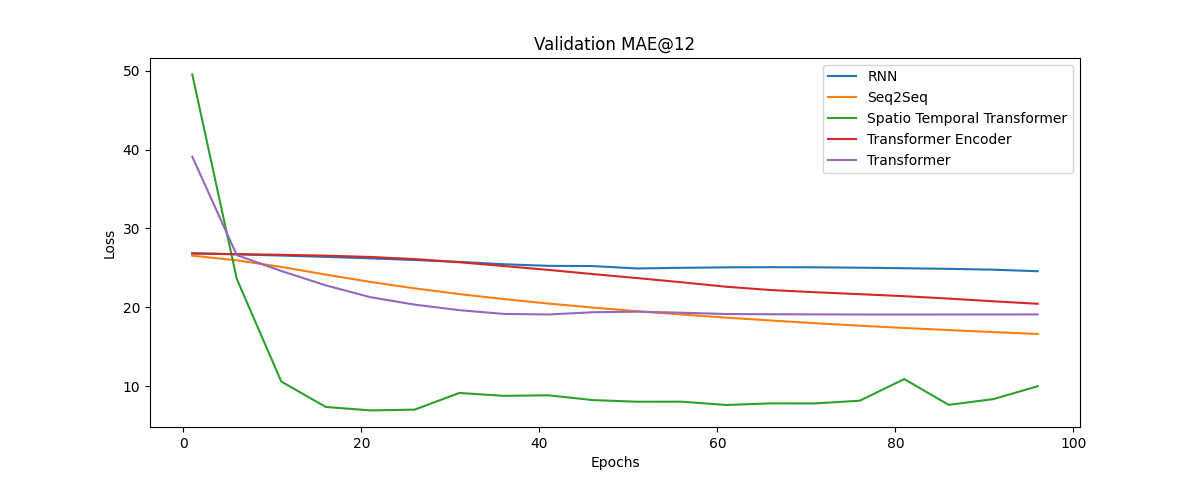}
\caption{\textbf Validation MAE@12}
\includegraphics[width=0.75\textwidth]{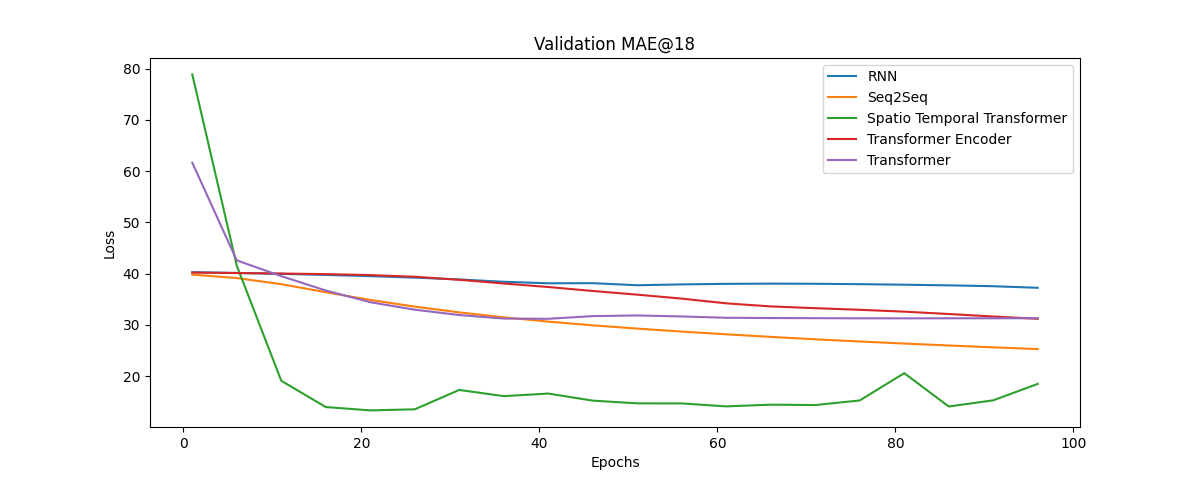}
\caption{\textbf Validation MAE@18}
\includegraphics[width=0.75\textwidth]{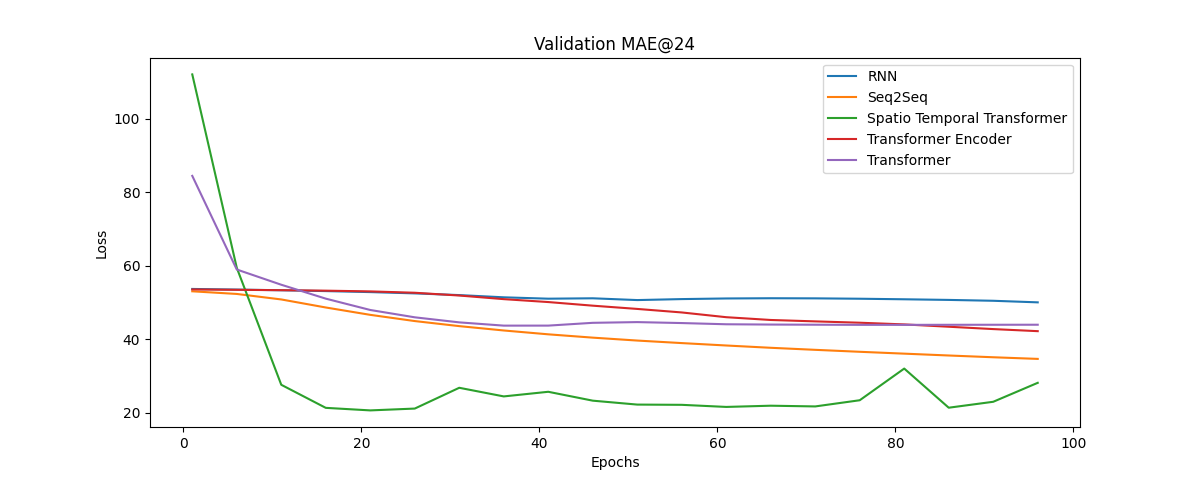}
\caption{\textbf Validation MAE@24}
\end{figure*}

\begin{figure*}[h]
    \centering
    \subfloat[\centering Frame 124]{{\includegraphics[width=2.5cm]{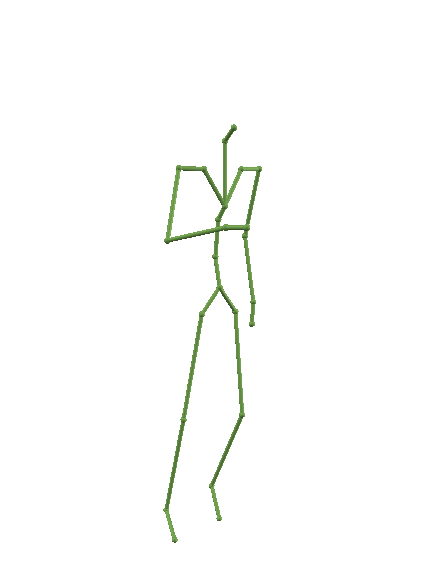} }}%
    \qquad
    \subfloat[\centering Frame 128]{{\includegraphics[width=2.5cm]{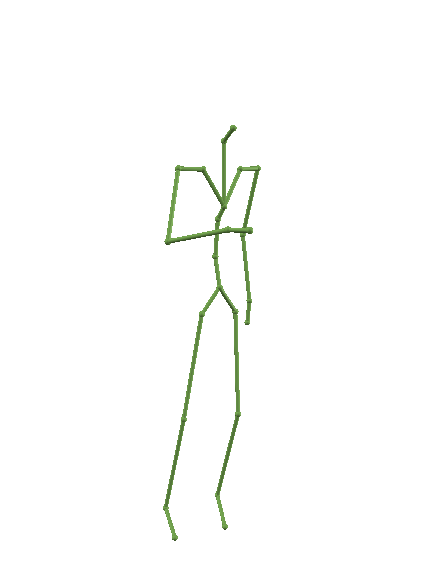} }}%
    \qquad
    \subfloat[\centering Frame 132]{{\includegraphics[width=2.5cm]{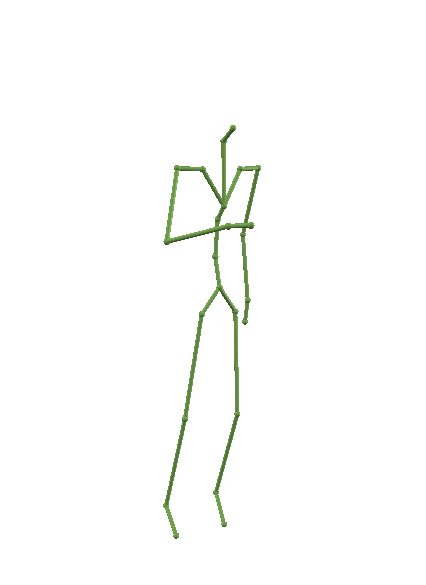} }}%
    \qquad
    \subfloat[\centering Frame 136]{{\includegraphics[width=2.5cm]{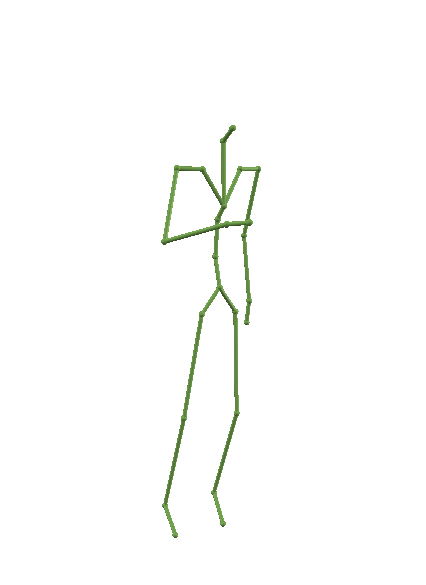} }}%
    \qquad
    \subfloat[\centering Frame 140]{{\includegraphics[width=2.5cm]{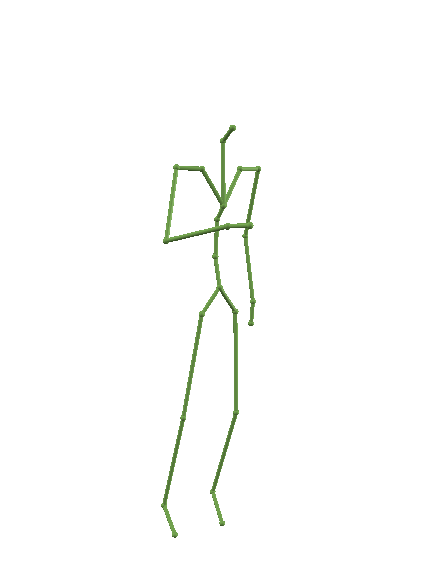} }}%
    \caption{Ground Truth Motion 1}%
    \label{fig:gt1}%
\end{figure*}
\begin{figure*}[h]
    \centering
    \subfloat[\centering Frame 124]{{\includegraphics[width=2.5cm]{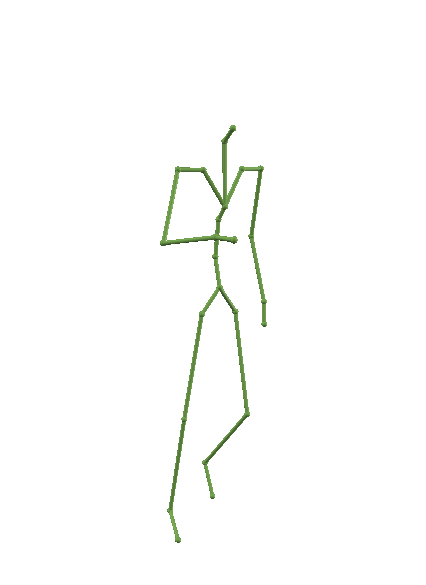} }}%
    \qquad
    \subfloat[\centering Frame 128]{{\includegraphics[width=2.5cm]{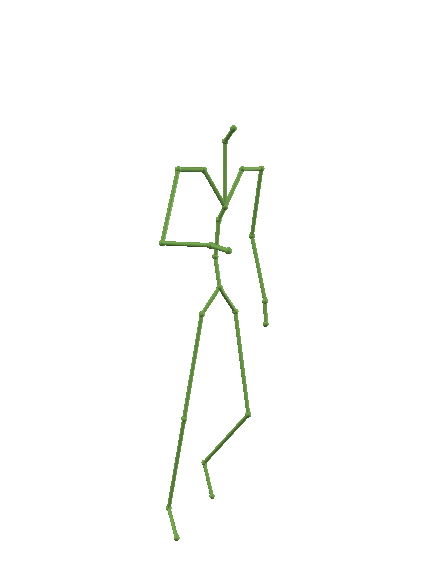} }}%
    \qquad
    \subfloat[\centering Frame 132]{{\includegraphics[width=2.5cm]{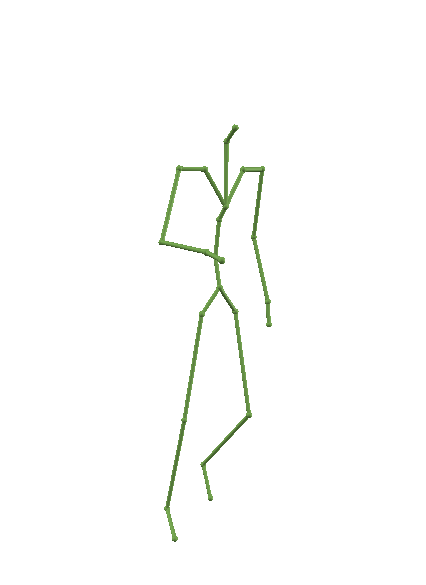} }}%
    \qquad
    \subfloat[\centering Frame 136]{{\includegraphics[width=2.5cm]{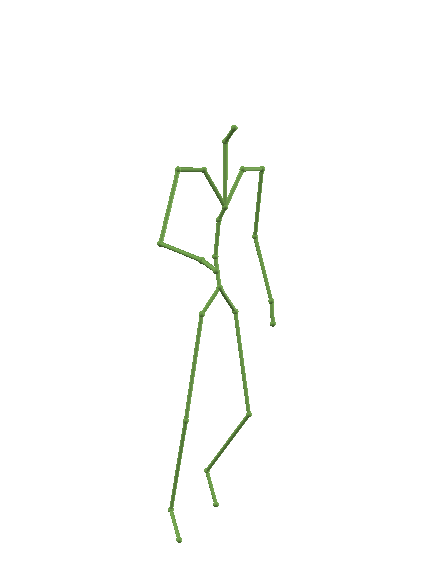} }}%
    \qquad
    \subfloat[\centering Frame 140]{{\includegraphics[width=2.5cm]{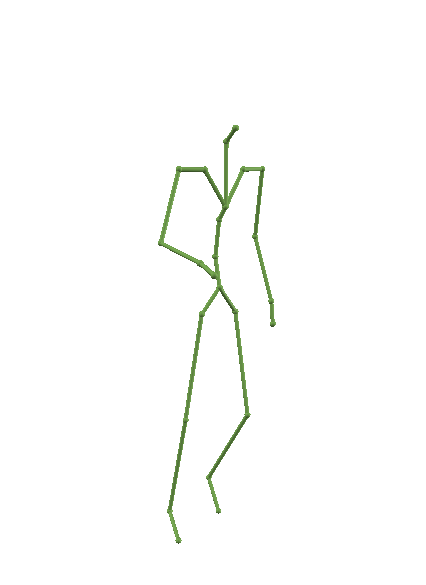} }}%
    \caption{ST Transformer Predicted Motion}%
    \label{fig:pm1}%
\end{figure*}
\begin{figure*}[h]
    \centering
    \subfloat[\centering Frame 124]{{\includegraphics[width=2.5cm]{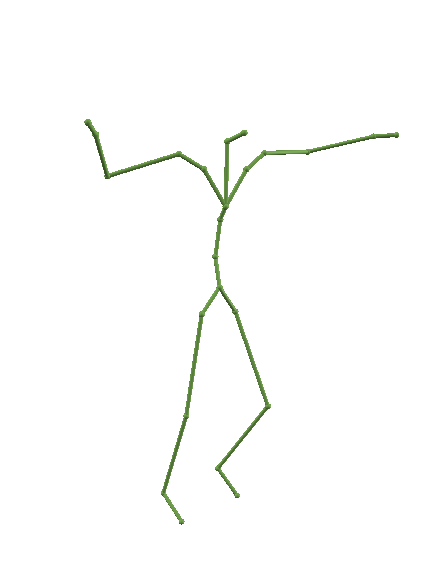} }}%
    \qquad
    \subfloat[\centering Frame 128]{{\includegraphics[width=2.5cm]{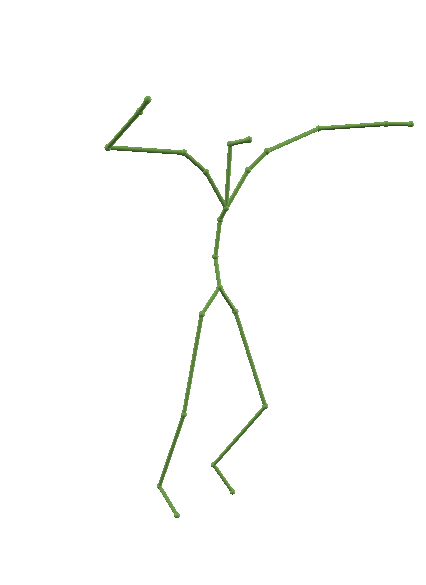} }}%
    \qquad
    \subfloat[\centering Frame 132]{{\includegraphics[width=2.5cm]{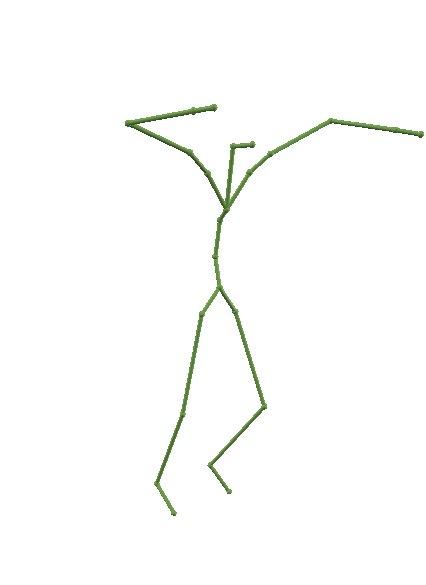} }}%
    \qquad
    \subfloat[\centering Frame 136]{{\includegraphics[width=2.5cm]{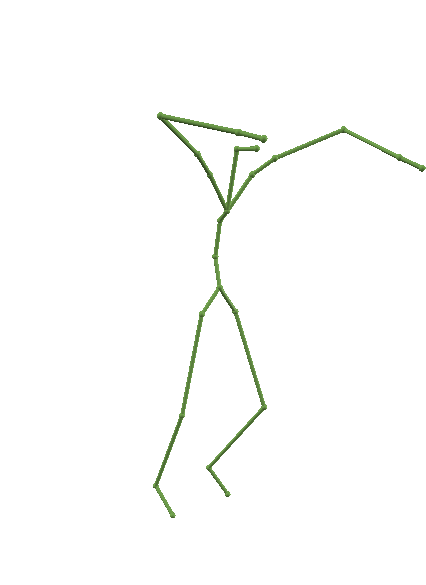} }}%
    \qquad
    \subfloat[\centering Frame 140]{{\includegraphics[width=2.5cm]{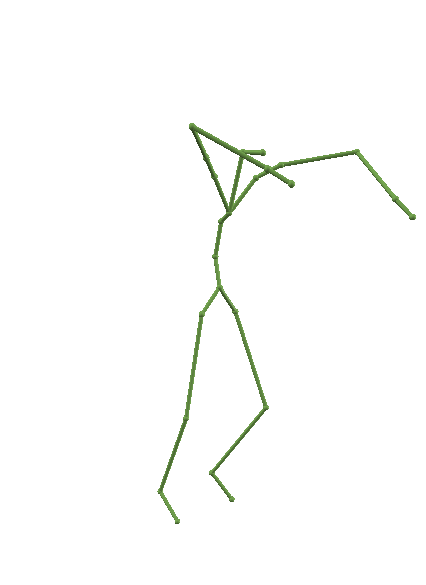} }}%
    \caption{Ground Truth Motion 2}%
    \label{fig:gt2}%
\end{figure*}
\begin{figure*}[h]
    \centering
    \subfloat[\centering Frame 124]{{\includegraphics[width=2.5cm]{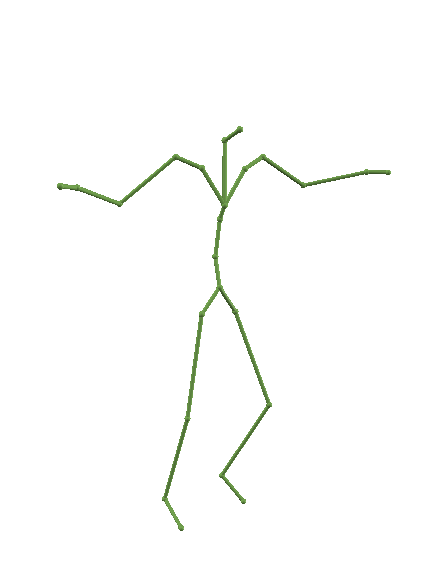} }}%
    \qquad
    \subfloat[\centering Frame 128]{{\includegraphics[width=2.5cm]{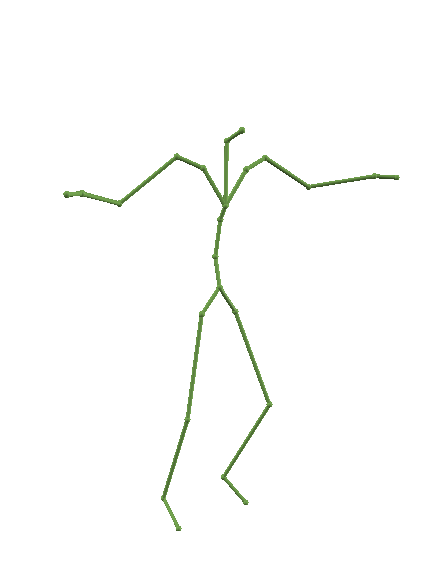} }}%
    \qquad
    \subfloat[\centering Frame 132]{{\includegraphics[width=2.5cm]{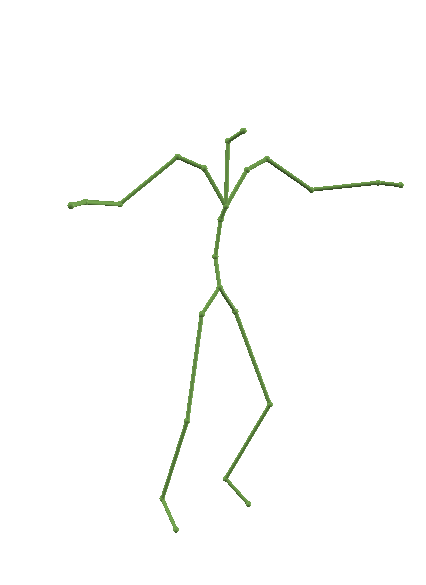} }}%
    \qquad
    \subfloat[\centering Frame 136]{{\includegraphics[width=2.5cm]{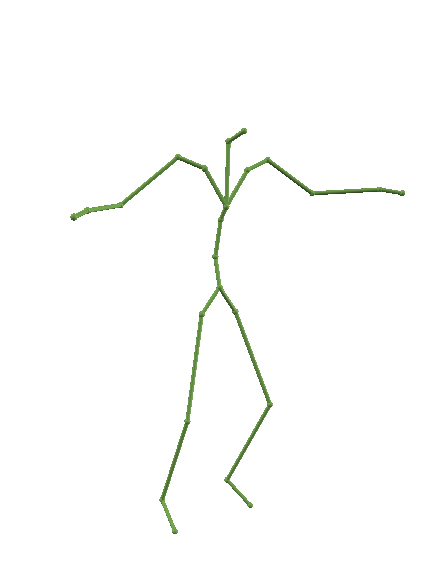} }}%
    \qquad
    \subfloat[\centering Frame 140]{{\includegraphics[width=2.5cm]{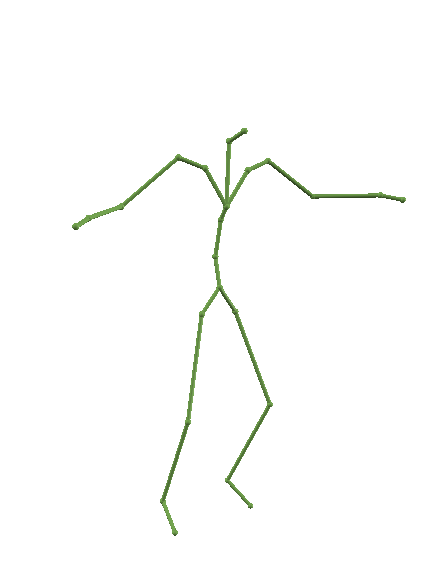} }}%
    \caption{MoE Predicted Motion}%
    \label{fig:pm2}%
\end{figure*}

\clearpage
\begin{table}[H]
\begin{center}
    \begin{tabular}{|c|c|c|c|c|}
         \hline
         \multicolumn{5}{|c|}{RNN, Seq2Seq, Transformer Encoder, Transformer, ST Transformer, and MoE performance after 100 epochs} \\
         \hline
         Architecture          & MAE@6         & MAE@12        & MAE@18        & MAE@24\\
         \hline
        RNN                   & 11.5310       & 23.2960       & 35.2423       & 47.3383       \\
        Seq2Seq               & 7.7943        & 15.1377       & 23.0770       & 31.8828       \\
        Transformer Encoder   & 9.2491        & 18.8478       & 28.7915       & 39.0187       \\
        Transformer           & 6.8229        & 16.9269       & 27.7262       & 38.9774       \\
        STTransformer         & 2.0733        & 6.1653        & 11.7492       & 18.2334       \\
        MoE                   & 1.9475        & 5.9395        & 11.3594       & 17.6420              \\
         \hline
    \end{tabular}
    \caption{Test MAE for best models}
    \label{tab:test_model_performance}
\end{center}
\end{table}
\clearpage
\section{Temporal Attention Visualizations}

\begin{figure}[h] 
  \centering
  \includegraphics[width=0.5\textwidth]{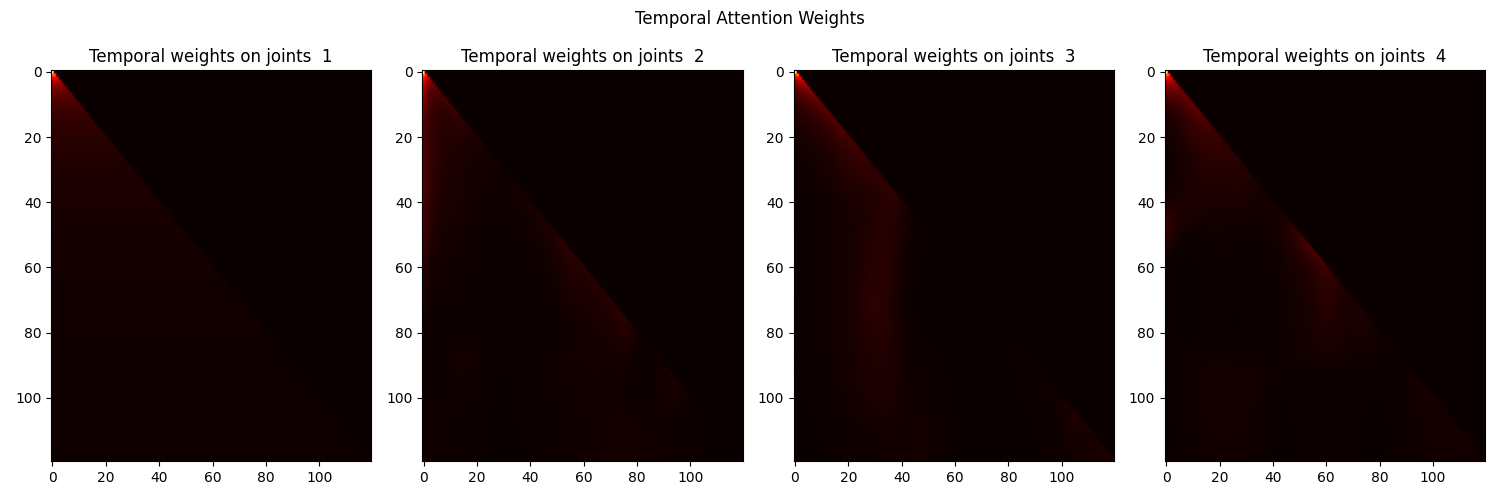} % ensure the filename matches and specify the desired width
  \caption{\textbf{Temporal Attention Map for joints 1, 2, 3, 4} map of temporal attention weights by timesteps in joints 1, 2, 3, 4}
  \label{fig:temporal_attention_1_2_3_4}
  \centering
  \includegraphics[width=0.5\textwidth]{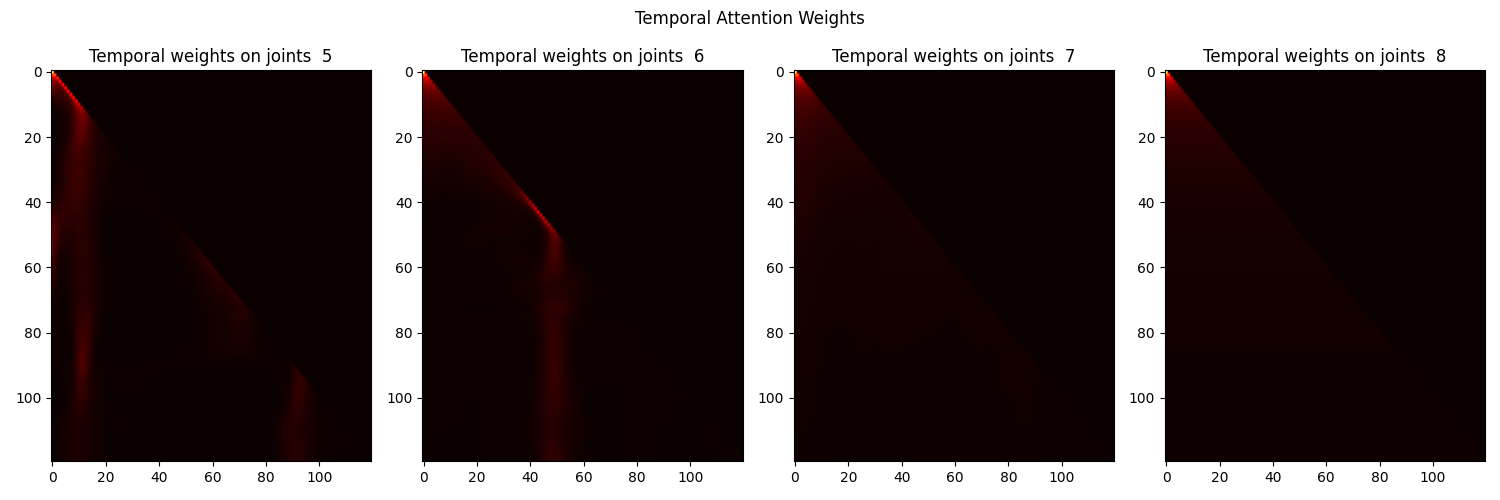} % ensure the filename matches and specify the desired width
  \caption{\textbf{Temporal Attention Map for joints 5, 6, 7, 8} map of temporal attention weights by timesteps in joints 5, 6, 7, 8}
  \label{fig:temporal_attention_5_6_7_8}
  \includegraphics[width=0.5\textwidth]{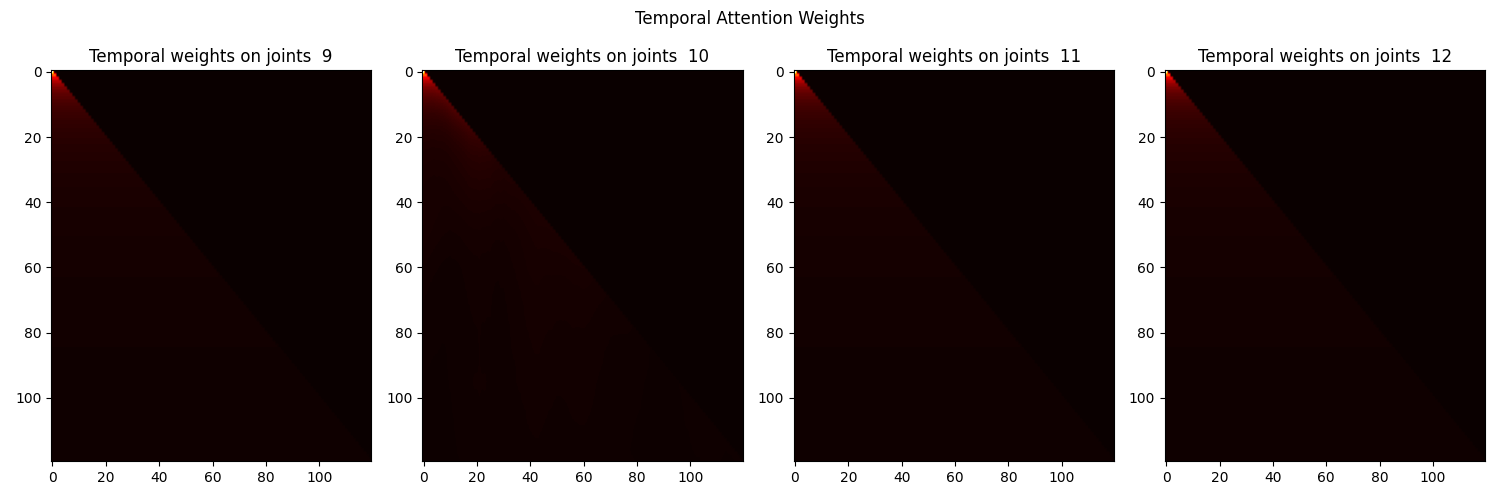} % ensure the filename matches and specify the desired width
  \caption{\textbf{Temporal Attention Map for joints 9, 10, 11, 12} map of temporal attention weights by timesteps in joints 9, 10, 11, 12}
\end{figure}
\vfill\eject

\begin{figure}[h] 
\vspace{0.43in}
  \label{fig:temporal_attention_9_10_11_12}
  \includegraphics[width=0.5\textwidth]{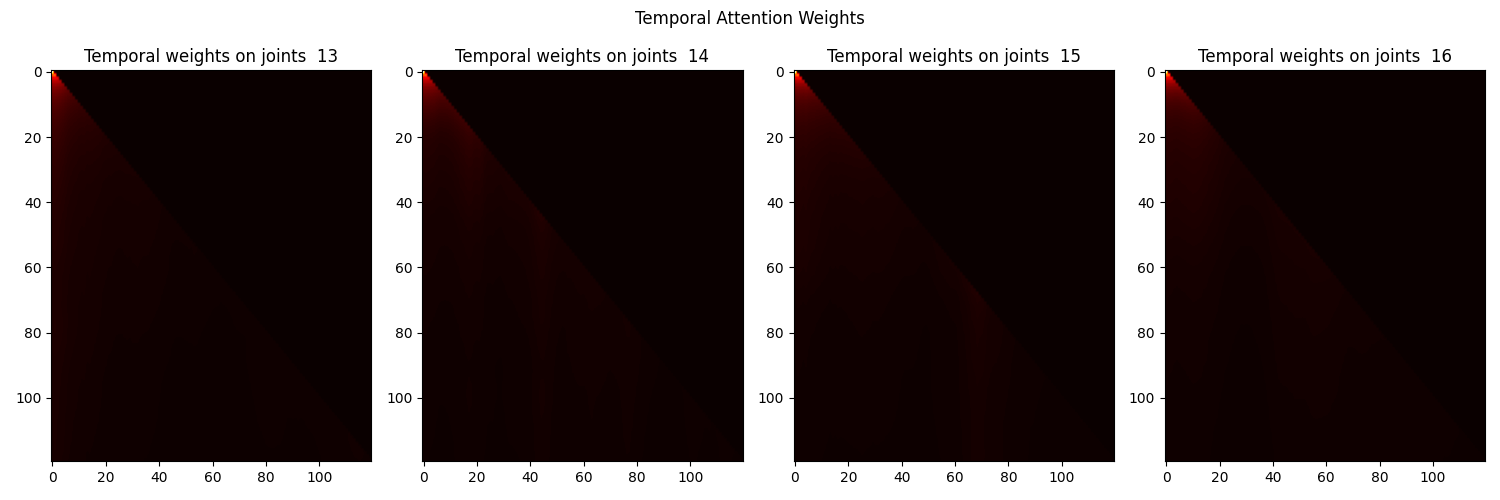} % ensure the filename matches and specify the desired width
  \caption{\textbf{Temporal Attention Map for joints 13, 14, 15, 16} map of temporal attention weights by timesteps in joints 13, 14, 15, 16}
  \label{fig:temporal_attention_13_14_15_16}
  \includegraphics[width=0.5\textwidth]{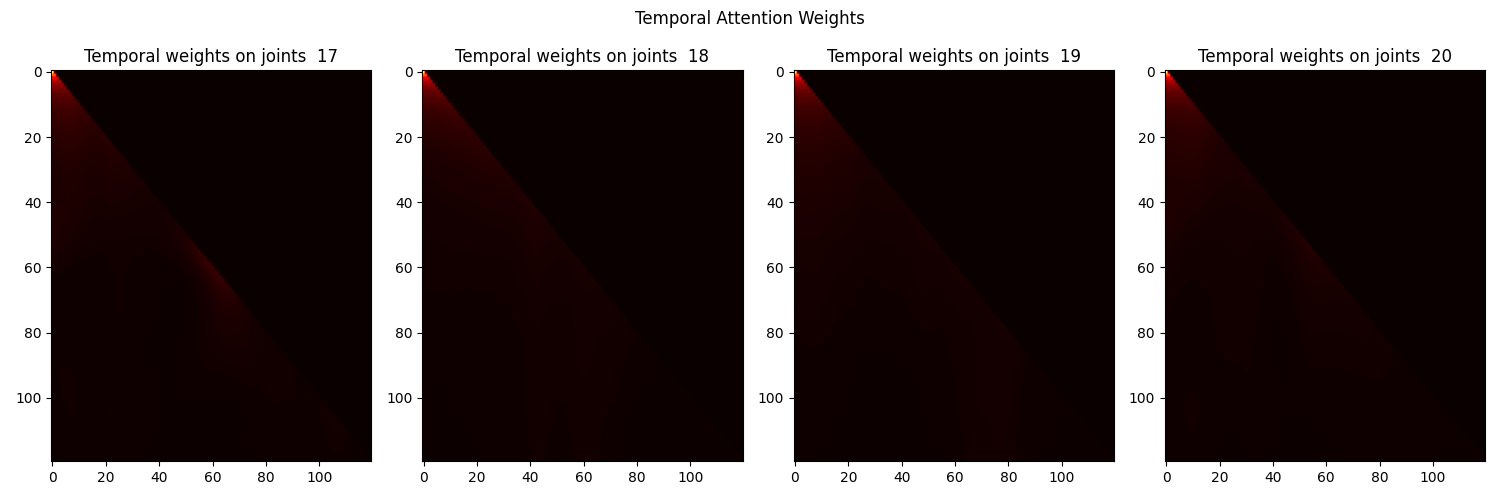} % ensure the filename matches and specify the desired width
  \caption{\textbf{Temporal Attention Map for joints 17, 18, 19, 20} map of temporal attention weights by timesteps in joints 17, 18, 19, 20}
  \label{fig:temporal_attention_17_18_19_20}
  \includegraphics[width=0.5\textwidth]{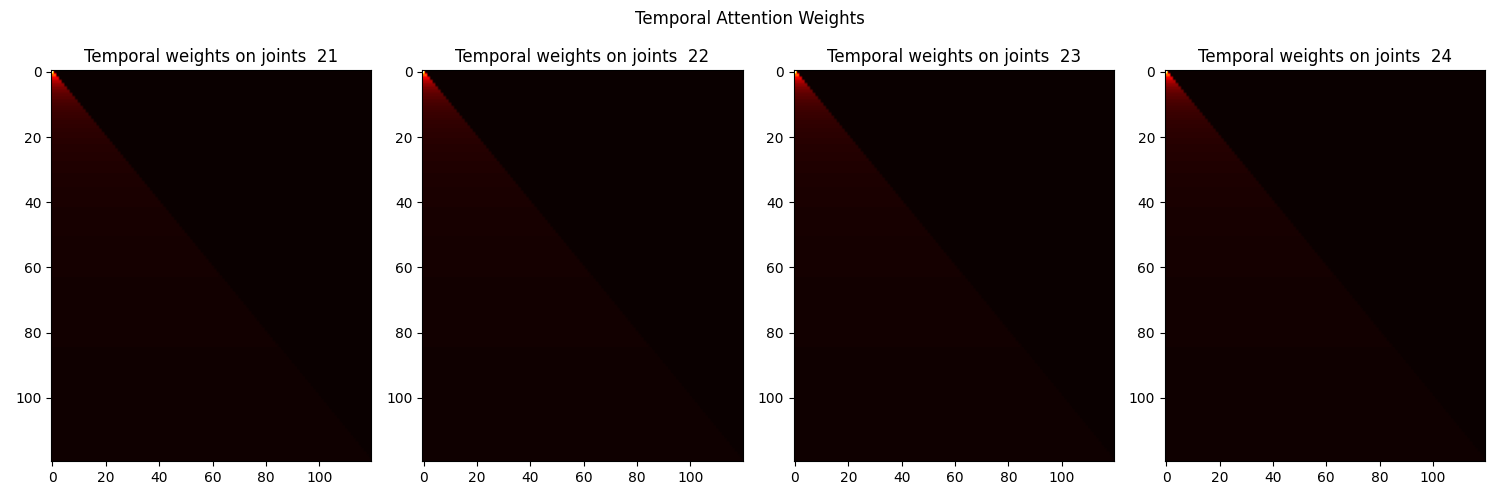} % ensure the filename matches and specify the desired width
  \caption{\textbf{Temporal Attention Map for joints 21, 22, 23, 24} map of temporal attention weights by timesteps in joints 21, 22, 23, 24}
  \label{fig:temporal_attention_21_22_23_24}
\end{figure}

\clearpage
\section{Loss Function}
\includegraphics[width=0.5\textwidth]{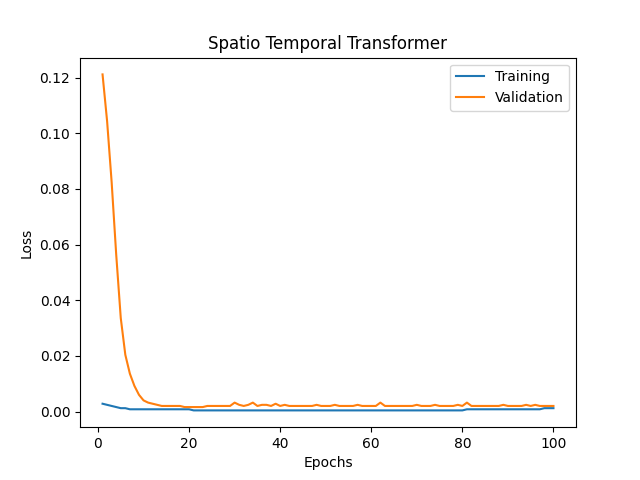}
\section{Inference Experiment Results}
\begin{table}[ht]
\centering
\caption{Testing Run Times for Different Model Configurations}
\begin{tabular}{|c|c|c|c|}
\hline
\textbf{Model} & \textbf{Param} & \textbf{Total Params} & \textbf{Testing Time (s)} \\ \hline
Vanilla  & 64 & 42.6K & 1.49 \\ \hline
Vanilla  & 128 & 50.8K & 1.54 \\ \hline
Vanilla  & 256 & 67.3K & 1.65 \\ \hline
Vanilla  & 512 & 100.3K & 1.78 \\ \hline
Vanilla  & 1024 & 166.4K & 2.34 \\ \hline
Vanilla  & 2048 & 298.5K & 3.38 \\ \hline
MoE  & 2 & 37.8M & 1.57 \\ \hline
MoE  & 4 & 75.6M & 1.62 \\ \hline
MoE  & 6 & 113.3M & 1.59 \\ \hline
MoE  & 8 & 151.1M & 1.74 \\ \hline
MoE  & 16 & 302.2M & 1.86 \\ \hline
MoE  & 32 & 604.3M & 2.35 \\ \hline
\end{tabular}
\label{tab:inference}
\caption{Param refers to hidden dimension of feedforward layer for STtransformer and number of experts for MoE.}
\end{table}

\end{document}